\documentclass[10pt,twocolumn,letterpaper]{article}

\usepackage{cvpr}              %

\usepackage{amsmath}
\usepackage{amssymb}
\usepackage{booktabs} %
\usepackage{multirow} %
\usepackage{caption}  %
\usepackage{adjustbox} %

\usepackage{natbib}

\definecolor{cvprblue}{rgb}{0.21,0.49,0.74}
\usepackage[pagebackref,breaklinks,colorlinks,allcolors=cvprblue]{hyperref}

\usepackage{url}

\usepackage{booktabs} %

\usepackage[ruled]{algorithm2e} %
\usepackage[capitalize]{cleveref}
\usepackage{graphicx}

\usepackage{array} %
\usepackage{amssymb}  %
\usepackage{amsfonts} %
\usepackage{bm} 
\usepackage{xcolor}   %
\usepackage[table]{xcolor}   %
\usepackage{pifont}

\usepackage{makecell} %
\usepackage{graphicx} %
\usepackage{multirow} %
\usepackage{colortbl}
\usepackage{placeins}

\usepackage{tcolorbox}
\usepackage{enumitem}

\usepackage{mathtools}

\usepackage{natbib}
\usepackage{float}

\newcommand{\name}{\textsc{CineScene}}

\def\paperID{} %
\def\confName{CVPR}
\def\confYear{2026}

\title{
\name: 
Implicit 3D as Effective Scene Representation for \\
Cinematic Video Generation
\vspace{-0.7em}
}

\author{
  \bf{Kaiyi Huang}\textsuperscript{$1^{\ast}$} \quad 
  \bf{Yukun Huang}\textsuperscript{$1$} \quad 
  \bf{Yu Li}\textsuperscript{$3$}  \quad 
  \bf{Jianhong Bai}\textsuperscript{$4$} \quad 
  \bf{Xintao Wang}\textsuperscript{$2{\dagger}$} \quad 
  \bf{Zinan Lin}\textsuperscript{$5$}  \\
  \bf{Xuefei Ning}\textsuperscript{$3$} \quad 
  \bf{Jiwen Yu}\textsuperscript{$1$} \quad 
  \bf{Pengfei Wan}\textsuperscript{$2$}  \quad 
  \bf{Yu Wang}\textsuperscript{$3$}  \quad 
  \bf{Xihui Liu}\textsuperscript{$1^{\dagger}$}  \\
\textsuperscript{1}The University of Hong Kong  \quad
\textsuperscript{2}Kling Team, Kuaishou Technology \quad 
\textsuperscript{3}Tsinghua University \\
\textsuperscript{4}Zhejiang University \quad 
\textsuperscript{5}Microsoft Research
\\
\href{https://karine-huang.github.io/CineScene/}{{\text{Project Page}}}
}

\begin{document}

\twocolumn[{%
\renewcommand\twocolumn[1][]{#1}%
\maketitle
\vspace{-3em}
\includegraphics[width=\linewidth]{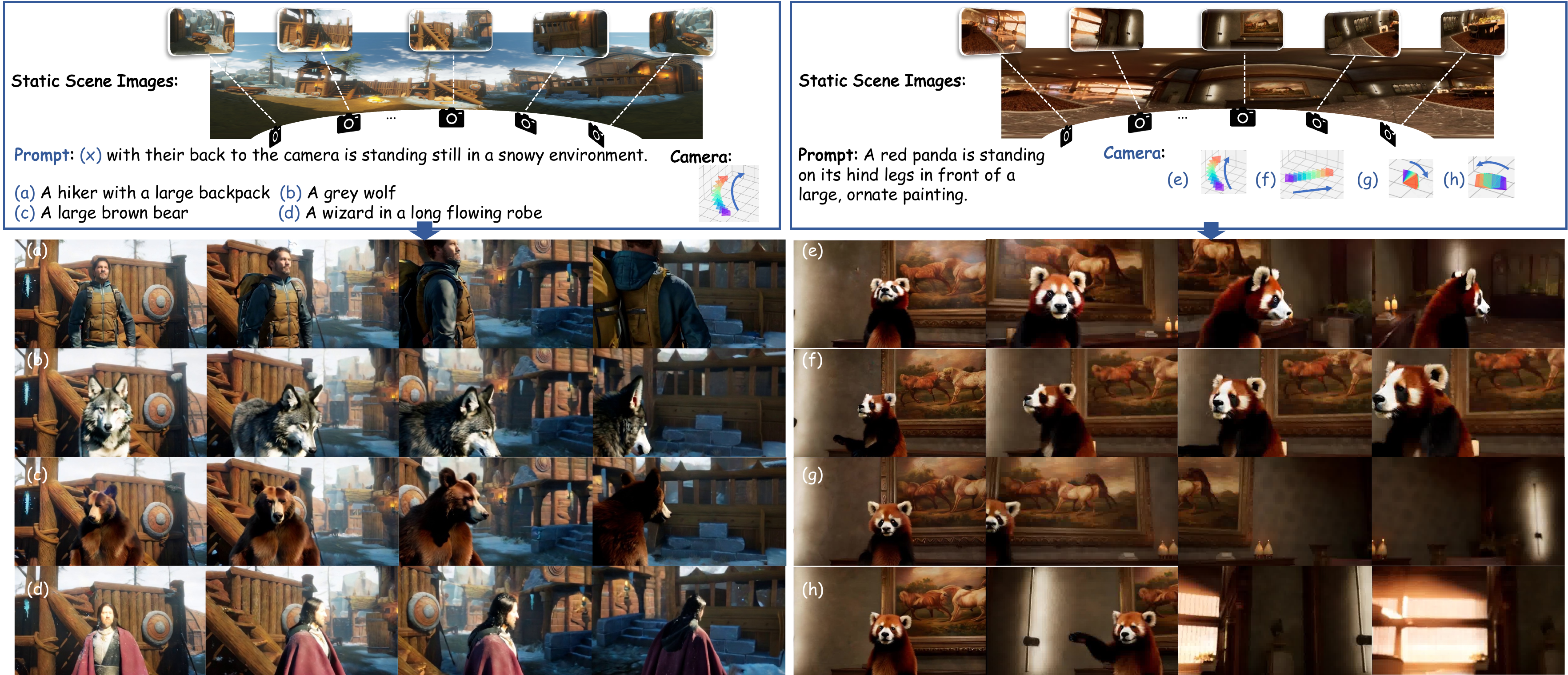}
\vspace{-2em}
\captionof{figure}{\textbf{Examples generated by \name{}}. Given multiple images of a static environment, prompt, and a user-defined camera trajectory, the model generates high-quality videos featuring dynamic subject, while preserving the underlying scene in large view changes.} 
\label{fig:teaser}

}]

\let\thefootnote\relax\footnotemark\footnotetext{$^{\ast}$Work done during an internship at Kling Team, Kuaishou Tech. $^{\dagger}$ Corresponding author.} 

\begin{abstract}
Cinematic video production requires control over scene-subject composition and camera movement, but live-action shooting remains costly due to the need for constructing physical sets. To address this, we introduce the task of cinematic video generation with decoupled scene context: given multiple images of a static environment, the goal is to synthesize high-quality videos featuring dynamic subject while preserving the underlying scene consistency and following a user-specified camera trajectory.
We present \name{}, a framework that leverages implicit 3D-aware scene representation for cinematic video generation. 
Our key innovation is a novel context conditioning mechanism that injects 3D-aware features in an implicit way: 
By encoding scene images into visual representations through VGGT, \name{} injects spatial priors into a pretrained text-to-video generation model by additional context concatenation, enabling camera-controlled video synthesis with consistent scenes and dynamic subjects. 
To further enhance the model's robustness, we introduce a simple yet effective random-shuffling strategy for the input scene images during training.
To address the lack of training data, we construct a scene-decoupled dataset with Unreal Engine 5, containing paired videos of scenes with and without dynamic subjects, panoramic images representing the underlying static scene, along with their camera trajectories. Experiments show that \name{} achieves state-of-the-art performance in scene-consistent cinematic video generation, handling large camera movements and demonstrating generalization across diverse environments.
Here is the \href{??}{Project Page}.
\end{abstract}
    
\section{Introduction}
\label{sec:intro}

Cinematic video production requires precise control over scene-subject composition and camera movement~\cite{mascelli1965fiveCofCinematography, christianson1996declarative}, making live-action shooting financially and logistically demanding~\cite{silva2024virtual}. 
In particular, constructing or modifying physical sets for different shots introduces repeated efforts and substantial costs. 
These challenges highlight the need for a more flexible framework that can create diverse visual narratives without extensive on-set resources. 
Recent advances in video generation~\cite{wan2025, hunyuanvideo2024community,kling} offer a promising avenue to create such a framework.
We therefore introduce the task of \textbf{cinematic video generation with decoupled scene context}. 
The goal is to generate a high-quality video featuring a new, dynamic subject based on three inputs: (1) a set of images defining a scene (serving as decoupled scene context), (2) a text prompt, and (3) a user-specified camera trajectory. The generated video must preserve the scene's consistency across large viewpoint changes while accurately following the desired camera motion, as shown in~\Cref{fig:teaser}.

Existing approaches for cinematic video generation face a fundamental trade-off between generative flexibility and scene consistency. 
2D context-based methods~\cite{guo2025lct, xiao2025captaincinema} operate directly in image space, offering high flexibility; however, they tend to struggle with maintaining scene consistency when large viewpoint changes occur, due to a lack of spatial understanding. In contrast, 3D-informed methods leverage depth cues~\cite{ren2025gen3c} or explicit 3D/4D reconstruction~\cite{bian2025gs, gu2025diffusion} to enforce scene consistency. Nevertheless, they tend to be complex and fundamentally limited by the challenge of obtaining accurate 3D/4D representations from sparse inputs~\cite{karaev2024cotracker, xiao2024spatialtracker}. The reliance on imperfect reconstructed geometry often hampers the overall quality of the generated output.

Recent 3D foundation models such as VGGT~\cite{wang2025vggt} have demonstrated the ability to obtain comprehensive spatial understanding from 2D images, represented by 3D-aware features. This suggests that robust 3D information can be obtained without resorting to explicit geometric reconstruction. Inspired by this, we propose \name{}, a cinematic video generation framework that leverages implicit 3D as scene representation to eliminate the need for explicit geometry.
The idea of integrating implicit 3D representations with video diffusion models has been recently explored. For instance, Geometry Forcing~\cite{wu2025geometryforcing} combines VGGT with a video diffusion model, using a VGGT-based loss to supervise the generation process. While effective for enforcing consistency, this loss-guided method is fundamentally limited to static scenes without new dynamic subject, as the supervision signal implicitly penalizes dynamic content (\Cref{fig:ablation_loss}).
\textit{Our key innovation} lies in how we integrate the implicit 3D. 
Rather than relying on a supervisory loss, \name{} introduces \textit{a context conditioning mechanism that directly integrates an implicit 3D scene representation into the video diffusion process}.
It enables the model to disentangle the static scene (provided as a conditioning input) from the dynamic content (to be generated), while jointly modeling a decoupled 3D structure and a dynamic subject. Consequently, our framework overcomes the limitation of prior work~\cite{wu2025geometryforcing, ren2025gen3c, yu2025contextasmemory}, which can only generate the static scene without new dynamic content, or are constrained to small view variations. In contrast, our approach generates scene-consistent videos with new dynamic subjects, even under large view changes.

Specifically, we encode static scene images into visual features through VGGT~\cite{wang2025vggt}, yielding an implicit 3D scene representation that captures both the spatial layout and camera information. These features are then projected and injected into a pretrained text-to-video (T2V) generation model as additional context tokens, allowing the model to preserve scene structure while generating vivid, dynamic content. To further strengthen the alignment between the scene images and their implicit 3D encoding, we introduce a simple but effective random-shuffling strategy for scene images during training. This strategy prevents the model from relying on fixed image ordering and encourages it to learn a robust correspondence between the generated content and the scene context.

Since no existing dataset explicitly separates static environments from dynamic subject, we construct a scene-decoupled dataset using Unreal Engine 5~\cite{unrealengine2022}. This dataset includes videos both with and without dynamic subject, panoramic images representing the underlying static scene, and corresponding camera trajectories. It provides the essential supervision needed to train scene-consistent generative models.

Our contributions are summarized as follows:
\begin{itemize}
    \item We introduce the task of cinematic video generation with decoupled scene context, enabling scene-consistent dynamic video synthesis in large view changes.
    \item To the best of our knowledge, we are the first to explore the context conditioning mechanism to inject implicit 3D scene representation that enables video generation models with spatial understanding. 
    \item We construct a scene-decoupled dataset containing panoramas, annotated camera trajectories, and paired videos with/without dynamic subject to enable explicit supervision of scene-consistent video generation with dynamic subject.
    \item Extensive experiments demonstrate that \name{} achieves state-of-the-art performance in cinematic video generation with strong scene consistency and camera accuracy, and exhibits out-of-domain generalization and practical applications in virtual stage production.
\end{itemize}

\section{Related Work}
\label{sec:related_work}

\subsection{Video Generation}
Video generation has progressed significantly recently, with diffusion model-based~\citep{ho2022imagen, singer2022make, zhou2022magicvideo, khachatryan2023text2video, luo2023videofusion, blattmann2023align, he2022latent, wang2023modelscope, yang2024cogvideox}, and 
language model-based~\citep{villegas2022phenaki, chang2023muse, kondratyuk2023videopoet, yu2023magvit, yu2023language, chang2022maskgit}. Video diffusion models excel by progressively refining noisy inputs into clean video samples, with recent advancements like Sora~\citep{videoworldsimulators2024}, HunyuanVideo~\citep{kong2024hunyuanvideo}, and Wan-Video~\citep{wan2025} demonstrating remarkably high-quality visual synthesis via sophisticated latent diffusion techniques. 
Video language models, such as VideoPoet~\citep{kondratyuk2023videopoet}, are typically derived from the family of transformer-based language models that can flexibly incorporate multiple tasks in pretraining, and show zero-shot capabilities.
Recent works~\citep{guo2025lct, xiao2025captaincinema, cai2025mixture}, such as Veo3~\citep{deepmind_veo3_2025}, RunwayML~\citep{gen4}, and Kling~\citep{kling} enable high-quality video generation.

\subsection{Context Conditioning}
Context conditioning paradigm for generation is pioneered in the image domain by OminiControl~\cite{tan2025ominicontrol}, and then extended to video by FullDiT~\cite{ju2025fulldit, he2025fulldit2}, and LCT~\cite{guo2025lct}. 
This simple yet effective design allows for flexible conditioning, but with only 2D contexts, it
still struggles with the challenge of scene consistency especially under large view changes due to the lack of spatial understanding.
Extensive works~\cite{wang2023incontext, xiao2025worldmem} utilizes a rule-based method to select overlap context images, however, they still need extra camera poses which are not estimated accurately. 
Recent works~\cite{ren2025gen3c, zhang2025worldconsistent} inject explicit 3D information (\textit{e.g.}, depth map, point clouds) into video model, however it is complex and hard to get accurate explicit 3D information as guidance.
For implicit guidance, prior works~\cite{wu2025geometryforcing, dai2025fantasyworld} investigate methods to enable models
to perceive 3D structural information within the diffusion process.
FantasyWorld~\cite{dai2025fantasyworld} introduces cross-branch supervision, where jointly generates videos and 3D attributes.
Geometry Forcing~\cite{wu2025geometryforcing} guides video model with 3D loss from VGGT~\cite{wang2025vggt}. However, this approach is limited to static scene generation, which restricts its applicability in cinematic video generation involving dynamic content.
Different from prior work, we explore the context conditioning mechanism to inject implicit 3D scene representation into a video model for generating novel videos, with no architecture modification. This allows creators to synthesize videos with dynamic content and flexible camera trajectories, all while maintaining scene consistency across large viewpoint changes.

\subsection{Cinematic Video Generation}
Cinematic video generation aims to incorporate cinematographic principles into both virtual and live-action videos~\citep{christianson1996declarative, kardan2008virtual, huang2025filmaster}. 
The Virtual Cinematographer (VC)~\citep{he2023virtualcinematographer, xu2025filmagent} introduces a paradigm for automatically generating complete camera specifications in real time for capturing events in 3D virtual environments. 
Other works focus on planning suitable camera placements~\citep{gleicher1992through, mackinlay1990rapid, phillips1992automatic, drucker1992cinema, drucker1995camdroid, drucker1994intelligent} or camera control~\cite{bahmani2024vd3d, yang2024direct, zheng2024cami2v, zheng2025vidcraft3, he2024cameractrl} for interactive tasks. 
While prior approaches primarily focus on camera setting strategies, our method considers scene consistency, dynamic content, and camera movements simultaneously, leveraging flexible novel cinematic video generation.

\section{Scene-Decoupled Video Dataset}
\label{sec: dataset}

\begin{figure}
    \centering
    \vspace{-10pt}
    \includegraphics[width=\linewidth]{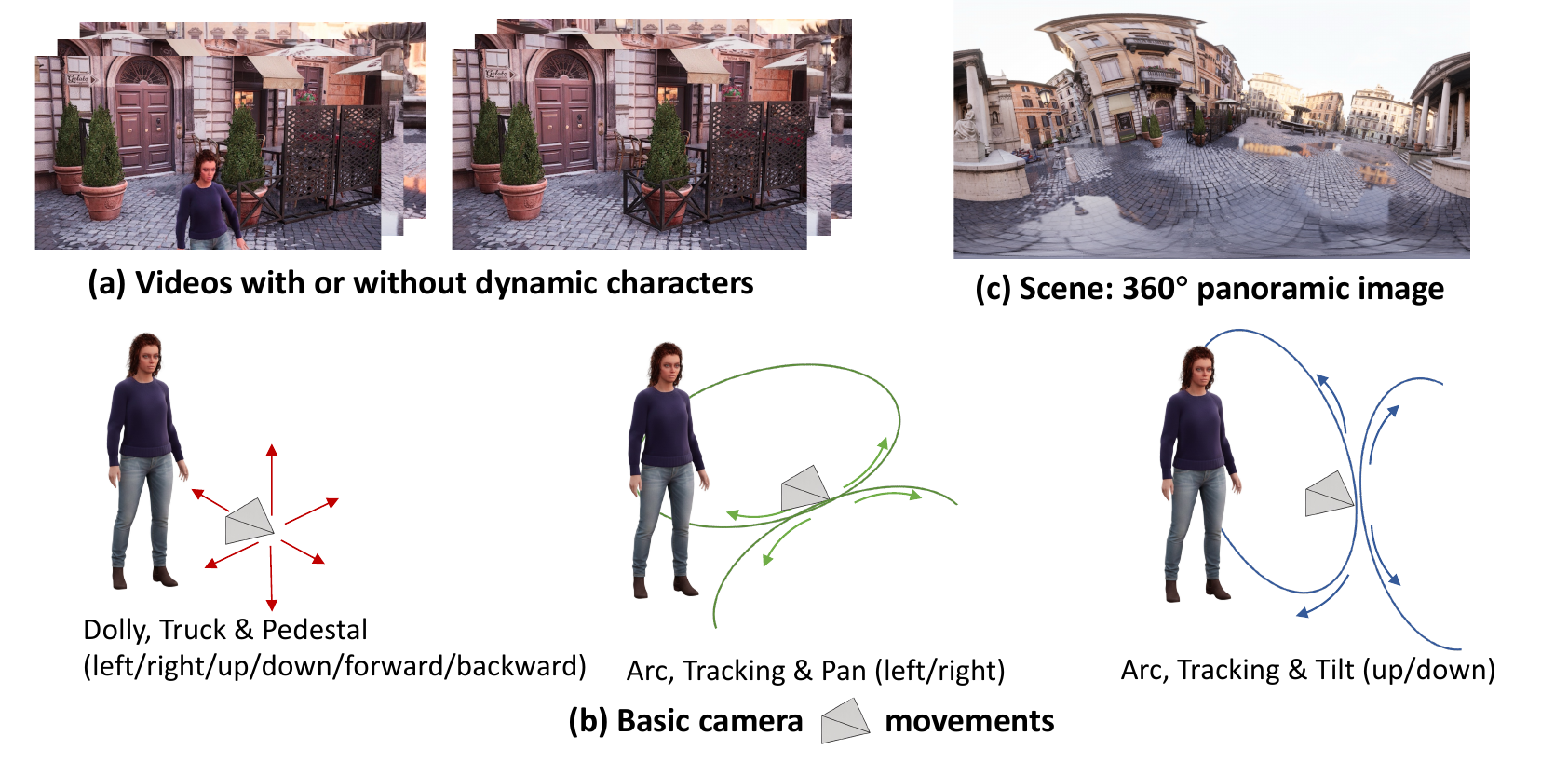}
    \vspace{-20pt}
    \caption{\textbf{Overview of Scene-Decoupled Video Dataset.} For each scene, we render (a) videos with/without dynamic subject, (b) 360° panoramic image representing the static scene from a common starting viewpoint, with (c) diverse camera trajectories. }
    \vspace{-20pt}
    \label{fig:dataset}
\end{figure}

To enable scene-consistent, camera-controlled video generation based on static scene images, data of videos with decoupled scene content and large view changes are required for training the video generation model. 
Specifically, the data should include videos with dynamic subjects and camera trajectories, paired with static scene images that are shot in the same location. 
Existing real-world datasets do not readily provide perfectly separated static backgrounds and dynamic foregrounds, nor do they offer the precise and diverse camera trajectory information essential for our task. Therefore, we use a rendering engine Unreal Engine 5~\cite{unrealengine2022} to generate the data. 
The advantages of this data collection pipeline are: 
1) It provides perfectly aligned pairs of dynamic videos and a static representation of the same scene. This perfect pairing is crucial for training a model to learn a decoupled representation of dynamic subjects and the static environment.
2) It enables precise and customizable camera trajectories that support large view changes.

The dataset is shown in~\Cref{fig:dataset}. For video data, we construct two versions: videos with or without dynamic subject. For the version with dynamic subject, we select randomly the animated human character from our asset library, and place it within multiple 3D environments as in~\cite{bai2025recammaster}. We randomly combine different characters and their actions in different scenes.
For the version without animated characters, we only remove the characters from the scene, leaving others unchanged.
For each scene, we define a set of camera trajectories originating from a common starting viewpoint. These trajectories are designed based on fundamental cinematographic movements~\cite{mascelli1965fiveCofCinematography}, including ``dolly'', ``pan'', ``tilt'', ``tracking'' along 6 directions, \textit{i.e.}, up, down, left, right, forward, backward. In order to achieve large view changes, we add constraints to the camera movements, such as 75 degrees for 77 frames in ``pan'' movement. More details can be found in Appendix. 
We randomly select a set of trajectories from basic camera movements for each location, all of which share a common starting viewpoint.
To represent the static scene for each location, we render a full 360-degree panoramic image from the shared starting viewpoint of the video trajectories. This panoramic image is then projected onto an equirectangular representation~\cite{huang2025dreamcube} to represent the static scene, where the horizontal axis
represents the longitude and the vertical axis represents the
latitude. This format is one of the commonly used for storing 360-degree scene content.
In total, we collect 46K video-scene image pairs in 35 high-quality 3D environments with 46K different camera trajectories.

\begin{figure*}
    \centering
    \vspace{-10pt}
    \includegraphics[width=\linewidth]{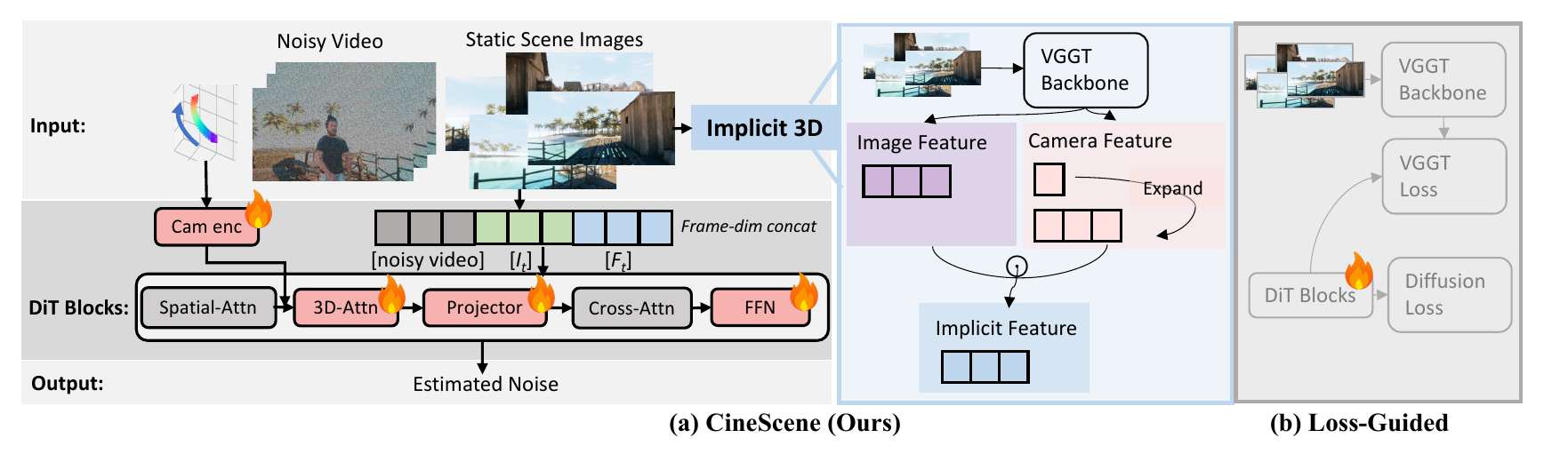}
    \vspace{-25pt}
    \caption{\textbf{Overview of \name{}.} \textit{Left}: Our method, \name{}, injects implicit 3D information as a context condition. Features from VGGT are encoded as tokens ($F_t$) and concatenated with the scene images ($I_t$) and the noisy video latents. This architecture fundamentally decouples the static background (the condition) from the dynamic foreground (the generation target). \textit{Right}: In contrast, loss-guided approaches use the VGGT features to form a supervisory loss, which penalizes deviations from the static scene and thus discourages dynamic content generation. We omit the text prompt for simplicity.} 
    \vspace{-20pt}
    \label{fig:pipeline}
\end{figure*}

\section{Method}

Given a set of decoupled scene images $I \in \mathbb{R}^{\times h \times w \times c}$, a prompt $P$, and a camera trajectory $C \in \mathbb{R}^{f\times3\times4}$, our objective is to generate a novel video $V \in \mathbb{R}^{f \times c \times h \times w}$ with dynamic subject and specified camera trajectory, where $c$ represents number of channels, $h$ and $w$ are image height and width.
$V$ should share the same scene consistency with $I$. To achieve this,
we propose to inject scene images along with their implicit 3D scene representation as conditions via context conditioning into video model.
The overview is shown in~\Cref{fig:pipeline}.

\subsection{3D-Aware Scene Representation Extraction}
In order to obtain a 3D-aware scene representation, we first get static scene images from the Scene-Decoupled Video Dataset, and then extract the implicit 3D scene representation from these images for subsequent video generation.

\noindent
\textbf{Scene Context Images.}
To simulate the static scene images captured by users, we perform an equirectangular-to-perspective projection~\cite{pyequilib2021github} based on the panoramic image. Specifically, we generate 20 perspective images from each equirectangular panorama by sampling the viewpoint along the horizontal plane at $18^\circ$ increments. These decoupled scene images $I$ cover the full $360^\circ$ horizontal scene. A wide field-of-view (FoV) of $90^\circ$ is employed to ensure comprehensive scene context is captured in each perspective image.

\noindent
\textbf{Implicit 3D Scene Representation.}
To equip the generation model with the necessary spatial understanding for maintaining scene consistency, we extract implicit 3D scene representation from the static scene images $I$. Specifically, we leverage the transformer backbone of VGGT~\cite{wang2025vggt} to extract features, which has been demonstrated to encode rich spatial information that significantly enhance downstream tasks like point tracking. We obtain the internal features from its final layer. These features are naturally decoupled into two components~\cite{wang2025vggt}: the image feature and the camera feature. The image feature $F_i \in \mathbb{R}^{20 \times k \times 2048}$ contains rich spatial cues, including information related to depth maps, point cloud structures, and tracking features. The camera feature $F_c \in \mathbb{R}^{20 \times 1 \times 2048}$, contains camera pose information. 
To enhance the generation model's holistic understanding of the 3D scene, we construct the final implicit features $F \in \mathbb{R}^{20 \times k \times 2048}$ by incorporating $F_i$ and $F_c$. $k$ is the number of tokens depending on the resolution. The incorporation is achieved by first expanding $F_c$ to match the spatial dimension of $F_i$ and then performing an element-wise addition. 
This fusion process effectively combines the scene's content information $F_i$ with its corresponding camera-viewpoint information $F_c$, creating a rich implicit representation for the video generation. We found this simple operation to be effective and efficient.

\subsection{Video Generation with Decoupled Scene Representation}
For video generation, we use scene context images, implicit 3D scene representation, camera, and prompt as inputs. Scene context images and implicit 3D scene representation are projected and injected via context conditioning before transformer blocks. 

\noindent
\textbf{Scene Context Images Condition.}
Given scene context images $I$, \name{} uses a causal 3D VAE~\cite{kingma2013vae} with temporal compression rate of 4 and spatial compression rate of 8 to encode images. Each image in $I$ ($\in \mathbb{R}^{h \times w \times 3}$) is first encoded to latent $\in \mathbb{R}^{h/8 \times w/8 \times 8}$ with VAE separately, then patchified with a patch size of 2 to get the image token $\in \mathbb{R}^{h/16 \times w/16 \times d}$, where $d$ represents the hidden dimension. Then all the image tokens are concatenated into a sequence in frame dimension as context tokens $I_{t} \in \mathbb{R}^{20 \times h/16 \times w/16 \times d}$. 

\noindent
\textbf{Implicit 3D Scene Representaion Condition.}
The implicit 3D scene is represented as implicit features $F \in \mathbb{R}^{20 \times k \times 2048}$, is first reshaped and interpolated in spatial dimension $\in \mathbb{R}^{20 \times h/8 \times w/8 \times 2048}$ for spatial alignment with $I$. Then it is patchified and projected to the hidden dimension with a convolutional layer and a layer norm to get implicit 3D tokens $F_{t} \in \mathbb{R}^{20 \times h/16 \times w/16 \times d}$. 
The input sequence for the transformer blocks is formed by concatenating the tokens of the noisy video, $I_{t}$, and $F_{t}$ along the frame dimension, allowing jointly modeling of scene images and implicit scene representation.

\noindent
\textbf{Camera and Prompt Condition.}
Unlike previous works~\cite{yu2025contextasmemory} that need explicit camera parameters from context images which is challenging to obtain accurate information even with the state-of-the-art structure-from-motion (SfM) methods~\cite{bai2025recammaster}, we only use the desired camera trajectory as input conditions. 
Given the input camera parameters $C \in \mathbb{R}^{f\times3\times4}$ that include the orientation and translation for each frame, we follow~\cite{bai2025recammaster} to apply camera injection to facilitate the model in correlating the camera parameters with the generated videos. 
Specifically, we project $C$ to have the same channels with the video tokens through a learnable camera encoder, and add it into the visual features that correspond to the noisy video. For the visual features that correspond to $I_{t}$, and $F_{t}$ tokens, we inject zero as a placeholder. 
We follow~\cite{bai2025recammaster} not to include camera intrinsics, as users are unlikely to have access to the source images' intrinsics. For simplicity in our experiments, we maintain a fixed relationship between the FoV of the context images and the generated video. However, our framework can be readily extended to incorporate varying camera intrinsics as an additional input condition with minimal modifications.
Then we inject the prompt condition into the cross attention of video model as prior work~\cite{rombach2022high}.

\noindent
\textbf{Discussion on Context Condition (Ours) \textit{v.s.} Loss-Guided Method.}
There has been exploration of integrating implicit 3D information into video diffusion models through a loss-guided mechanism~\cite{wu2025geometryforcing}. This approach fine-tunes the video model by minimizing the discrepancy between VGGT features and the diffusion model’s latent representations. While effective, our experimental results demonstrate that our context-conditioning strategy for injecting implicit 3D representations offers substantial advantages over the loss-guided approach, particularly for generating dynamic content (\Cref{fig:ablation_loss}).
We attribute this superior performance to two key aspects of our method. First, by injecting the scene representation as a context condition rather than as a supervisory loss, our framework inherently decouples the static background from the dynamic foreground. This provides a more robust mechanism for generating vivid motion without being constrained by a loss function optimized for static reconstruction. 
Second, our approach aligns better with the diffusion paradigm by using VGGT features as guiding context instead of as a separate training objective, unlike the loss-guided method where VGGT supervision and diffusion loss act independently. Our method enables the diffusion model to jointly model the decoupled scene and dynamic subjects.

\subsection{Shuffled Context Images Alignment}
\label{method: shuffled}

We observed that using ordered scene context images as inputs for training (which are at equal $18^\circ$ incremental degrees in the horizontal plane) will dominate the generation model with pixel information, especially the content of the first and last images (\Cref{fig:ablation_shuffle}). This leads the generation model neglect to learn from the implicit 3D scene representation. 
Possible solution includes progressive training strategy~\cite{ju2025fulldit}, that first trains difficult-to-learn part early to ensure the model learn robust representation, and then introduces easier part that benefits from improved feature representation.
However, progressive training (train first implicit condition, and then add scene context images) is not suitable for our task, hindering the joint alignment of scene context images and their implicit 3D representation.
We attribute this observed issue to position-aware prior introduced by the positional embedding~\cite{su2024roformer} within the video generation model. To mitigate this issue, we propose the context image shuffling mechanism during training. Specifically, we fix the position of the first context image (which corresponds to the starting viewpoint of the target video) in the input sequence, while randomly shuffling the order of the remaining context images.
We found that this simple yet effective mechanism enables the generation model to learn the alignment between pixel-level context and implicit 3D scene representation, rather than exploiting correlations from a fixed input order.

\begin{table*}[htbp] %
    \centering
    \vspace{-10pt}
    \caption{\textbf{Quantitative comparision with previous methods.} We compare \name{} with FramePack~\cite{zhang2025framepack} on scene consistency, Context-as-Memory~\cite{yu2025contextasmemory} and Gen3C~\cite{ren2025gen3c} on both scene consistency and camera accuracy, Traj-Attn~\cite{xiao2025trajectory} and RecamMaster~\cite{bai2025recammaster} on camera accuracy. We follow~\cite{xie2025simplegvr} to evaluate video quality on VBench.
    }
    \vspace{-10pt}
    \label{tab:compare_baselines} %
    \adjustbox{max width=\textwidth}{
    \begin{tabular}{l ccc ccc ccc c cc} %
        \toprule
        \multirow{2}{*}{\textbf{Model}} & \multicolumn{5}{c}{\textbf{Scene Consistency}} & \multicolumn{3}{c}{\textbf{Camera Accuracy}} & {\textbf{Text Alignment}}  & {\textbf{Video Quality}}
        \\
        \cmidrule(lr){2-6} \cmidrule(lr){7-9} \cmidrule(lr){10-10} \cmidrule(lr){11-11}
        & Mat. Pix.(K)$\uparrow$  & CLIP-V$\uparrow$ & PSNR$\uparrow$ & SSIM$\uparrow$ & LPIPS$\downarrow$ & RotErr$\downarrow$ & TransErr$\downarrow$ & CamMC$\downarrow$ & CLIP-T$\uparrow$  & VBench$\uparrow$ \\
        \midrule
        \textbf{Context-based Method} \\ %
        \midrule
        FramePack~\cite{zhang2025framepack} &4107.45	&0.8421	&11.8854	&0.3551	&0.5505	&-	&-	&-	&\textbf{0.3269}  & 0.7999\\
        Context-as-Memory~\cite{yu2025contextasmemory} &4581.15	&0.8542	&13.8102	&0.3981	&0.4486	&2.7106	&5.2194	&6.9554	&0.3199   & 0.8018	\\
        \midrule
        \textbf{Explicit 3D Guidance Method} \\ %
        \midrule
        Gen3C~\cite{ren2025gen3c} &4541.25	&	0.8292	&11.6255	&0.3125	&0.6711	&2.9670	&10.1578	&11.1018	&0.2824 & 0.7585\\
        \midrule
        \textbf{Camera-Controlled Method} \\ %
        \midrule
        Traj-Attn~\cite{xiao2025trajectory} & - & - & - & - & - & 4.7124 & 7.8844 & 11.0607 & 0.2653  & 0.7771\\
        RecamMaster~\cite{bai2025recammaster} & - & - & - & - & - & 3.0854 & 7.3714 & 9.2358 & 0.3086 & 	0.7950\\
        \midrule
        \textbf{Ours} & \textbf{4617.51}  & \textbf{0.8633} & \textbf{14.5094} & \textbf{0.4133} & \textbf{0.4241} &\textbf{2.6825}	&\textbf{5.1460}	&\textbf{6.8819} &  \underline{0.3212}  & \textbf{0.8053}	\\ %
        \bottomrule
    \end{tabular}
    }
    \vspace{-10pt}
\end{table*}

\section{Experimental Results}

\subsection{Experimental Settings}
\textbf{Implementation Details.}
We train \name{} based on an internal text-to-video diffusion model with the Scene-Decoupled Video Dataset introduced in~\Cref{sec: dataset}. The model is trained for 10K steps with a batch size of 16, a learning rate of $5 \times 10^{-5}$, and a timestep shift of 15. 
We use 20 scene context images, target camera pose, and target prompt as inputs to the model.
For inference, we use a resolution of $384 \times 672$ with 77 frames. We set the inference step number as 50.

\noindent
\textbf{Baselines.}
Due to the lack of existing works with the identical experimental setting, we select the following methods as baselines for comparison: (1) \textit{Context-based method}: FramePack~\cite{zhang2025framepack}, Context-as-Memory (CaM)~\cite{yu2025contextasmemory}. 
FramePack hierarchically compresses context into frames. 
Context-as-Memory uses a rule-based FoV selection method, which needs the camera pose of context images. Here we use VGGT~\cite{wang2025vggt} to predict camera poses of context images.
(2)\textit{ Explicit 3D guidance method}: Gen3C~\cite{ren2025gen3c}, conditioning on the 2D renderings of predicted point clouds of images with camera trajectory.
(3) \textit{Camera-controlled method}: Trajectory-Attention (Traj-Attn)~\cite{xiao2025trajectory}, ReCamMaster~\cite{bai2025recammaster}. 

\noindent
\textbf{Evaluation.}
We employ 10 metrics across four key aspects: scene consistency, camera control, text alignment, and overall video quality.
(1) \textit{Scene consistency}: we follow prior work~\cite{bai2025recammaster, luo2025camclonemaster} to use GIM~\cite{shen2024gim} to calculate the number of matching pixels with confidence greater than the thresholdand (Mean Pixel Error (Mat. Pix.)), and CLIP-V~\cite{radford2021learning} (frame-wise CLIP similarity).
We also follow~\cite{ren2025gen3c} to calculate pixel-align metrics, PSNR and SSIM~\cite{wang2004ssim}, and perceptual metrics, LPIPS~\cite{zhang2018lpips}.
(2) \textit{Camera accuracy}: we adopt RotErr, TransErr, and CamMC as in CamI2V~\cite{zheng2024cami2v}. 
(3) \textit{Text alignement}: we use CLIP-T~\cite{radford2021learning}, following prior work~\cite{bai2025recammaster}.
(4) \textit{Video quality}: 
we evaluate on VBench~\cite{huang2023vbench} metrics.
For test set, due to the lack of existing decoupled scene content data, we follow~\cite{yu2025contextasmemory} to randomly choose 300 samples from the held out of
the Scene-Decoupled Video Dataset containing scenes with or without dynamic subject for testing. To test the out-of-domain ability, we construct 50 samples from DiT360~\cite{feng2025dit360, Matterport3D}, which are real-world scenes. We provide OOD qualitative results in~\Cref{exp: generalization}, and quantitative results in Appendix.

\begin{figure*}[htbp]
    \centering
    \vspace{-10pt}
    \includegraphics[width=\linewidth]{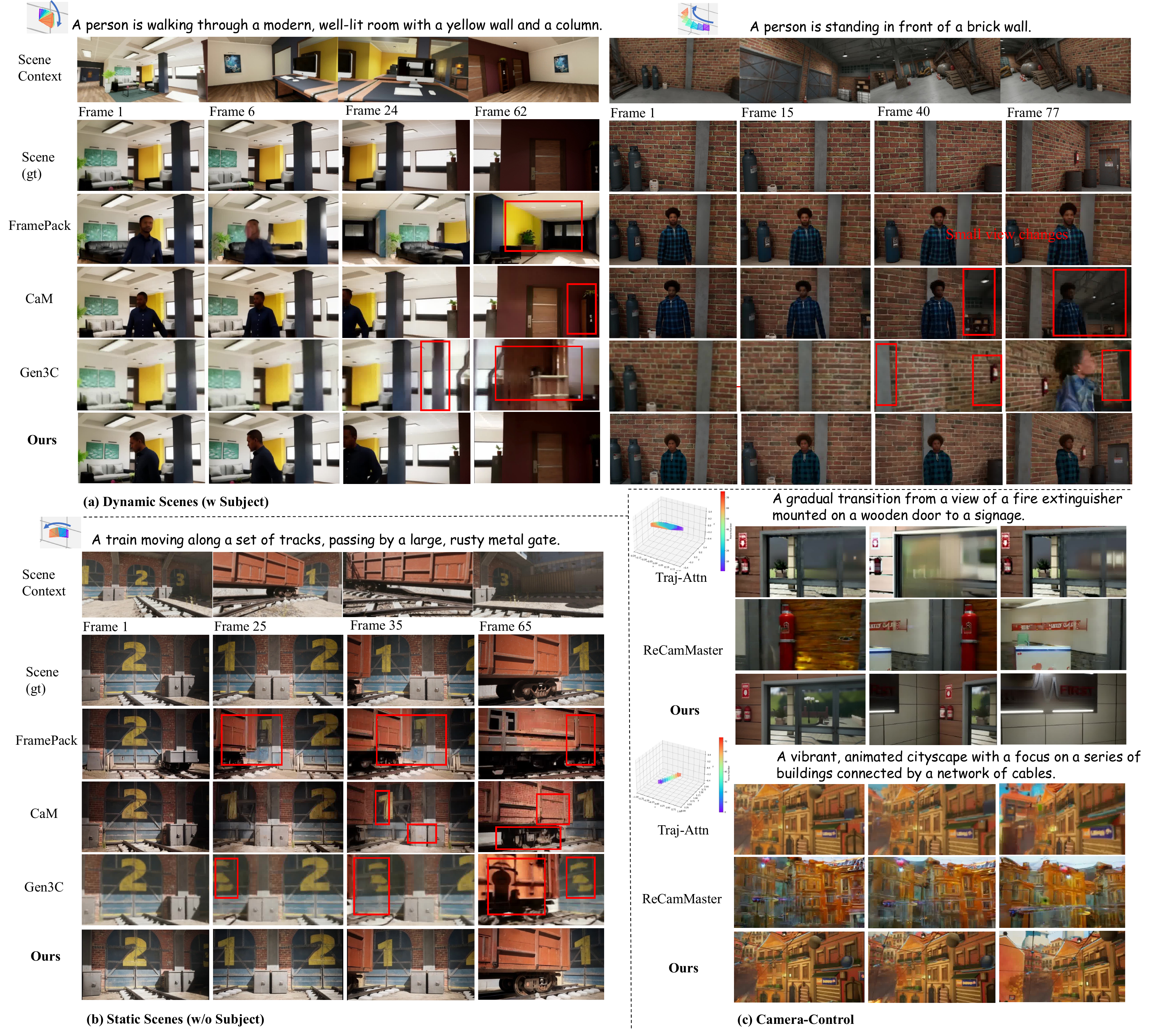}
    \vspace{-20pt}
    \caption{
    \textbf{Qualitative comparison of \name{} and previous context-based, explicit 3D guidance, camera-controlled methods.} We present dynamic scenes, static scenes compared with FramePack~\cite{zhang2025framepack}, CaM~\cite{yu2025contextasmemory}, and Gen3C~\cite{ren2025gen3c}, camera-control with Traj-Attn~\cite{xiao2025trajectory} and RecamMaster~\cite{bai2025recammaster}. We provide scene ground truth (gt) for comparison.
    We only show 4 scene context images for illustration. 
    }
    \vspace{-15pt}
    \label{fig:compare_all}
\end{figure*}

\subsection{Comparison with State-of-the-Art Methods}

\noindent
\textbf{Quantitative Results.}
For fair comparison, all context-based methods~\cite{zhang2025framepack, yu2025contextasmemory} are reimplemented on our base model and dataset with identical training setting. 
We employ the T2V model from~\cite{bai2025recammaster} as its I2V model is not open-source.
As show in~\Cref{tab:compare_baselines}, we evaluate scene consistency by comparing \name{} against both context-based and explicit 3D guidance methods. The quantitative metrics, including Mat. Pix., CLIP-V, PSNR, SSIM, and LPIPS, consistently demonstrate \name{}'s superior performance in maintaining scene consistency. This superiority stems from \name{}'s effective utilization of joint scene context images coupled with their 3D spatial understanding.
FramePack performs poorly in scene consistency due to its reliance on compressed and purely 2D pixel-level scene information. Context-as-Memory struggles with consistency, especially under large view changes, which we attribute to its dependency on both high-overlap context images, and accuracies in estimated camera poses. 
Gen3C also struggles with scene consistency as it uses explicit 3D guidance, and the inaccuracy of 3D guidance for large view changes causes inconsistency. Furthermore, the method that relies on explicit 3D guidance requires additional inference time, about 10.17× longer than ours.
In contrast, our method's use of implicit 3D information enables it to effectively align the generated video with the underlying 3D structure.
Then, we evaluate camera accuracy against methods that are either camera-controlled~\cite{bai2025recammaster, xiao2025trajectory} or use camera parameters as input ~\cite{yu2025contextasmemory, ren2025gen3c}. The results show that by leveraging implicit 3D information, our method also achieves higher camera accuracy. 
The text alignment of \name{} is slightly lower than that of FramePack since the calculation of text alignment is based on CLIP similarity between frames and the text. As \name{} exhibits significantly greater camera movement compared to FramePack, the text alignment metric is impacted by the large change between frames.

\noindent
\textbf{Qualitative Results.}
As illustrated in~\Cref{fig:compare_all}, \name{} demonstrates superior scene consistency under large view changes in both dynamic scenes and static scenes, and accuracy in camera control.
FramePack can generate novel videos but fails to generate scene-consistent videos with large view changes or limited to a fixed view.
Gen3C causes significant scene changes with large view changes, because it relies on the projection of the predicted pixel-wise depth of context images in large view change. When the predicted 3D guidance is not sufficient or accurate, it results in inconsistency.
Since context-as-memory and \name{} are trained on the same backbone, this highlights the effectiveness of scene implicit 3D information injection for scene consistency.

\subsection{Ablation Study}
\begin{figure}[t]
    \centering
    \includegraphics[width=\linewidth]{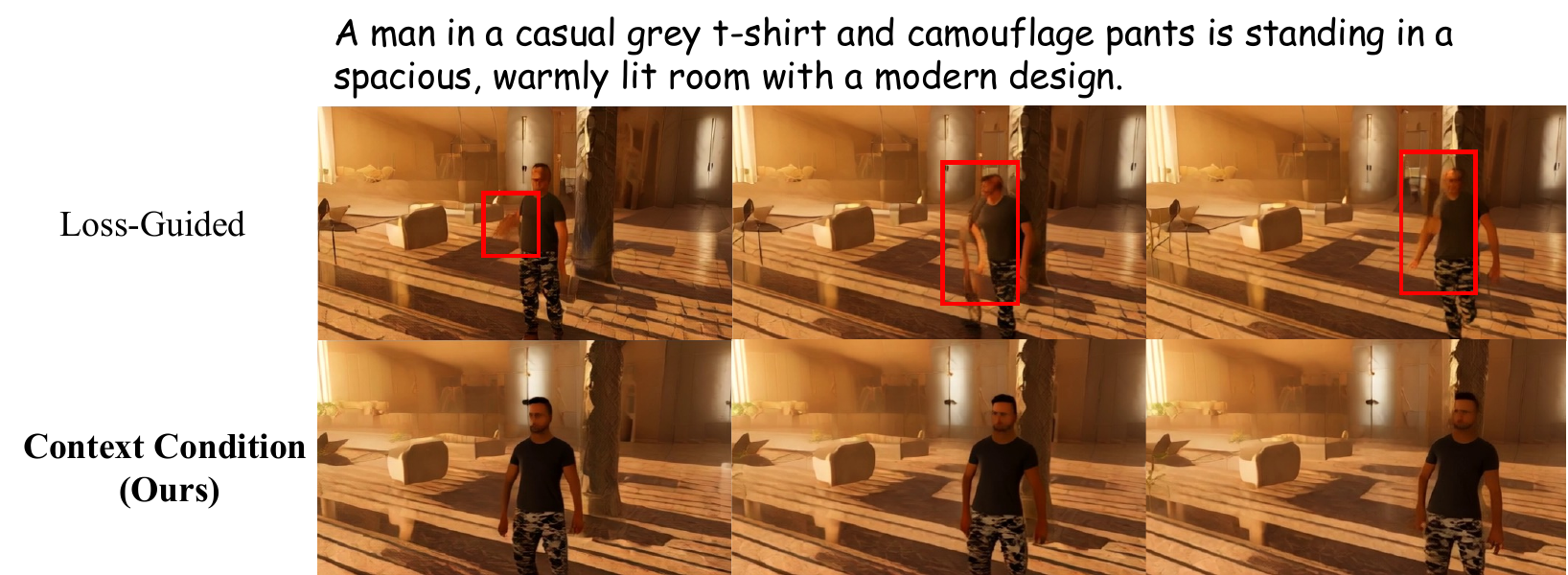}
    \vspace{-20pt}
    \caption{\textbf{Qualitative ablation study on injecting implicit 3D methods.} Loss-guided method shows artifacts when generating dynamic subject.
    }
    \vspace{-20pt}
    \label{fig:ablation_loss}
\end{figure}

\noindent
\textbf{Ablation on Different Implicit 3D Methods.} 
We compare our context conditioning (ours) with loss-guided method. The results in~\Cref{tab:ablation_vggt} show that the loss-guided method cause scene inconsistency compared to ours (Loss-Guided \textit{v.s.} Ours). Also we observed that loss-guided injection method suffer from significant artifacts of dynamic subject due to the penalized loss on dynamic content (\Cref{fig:ablation_loss}).

\noindent
\textbf{Impact of Scene Implicit 3D Representation.}
We train four sets of models with different implicit 3D information conditions: (1) without implicit 3D; (2) implicit 3D with only image feature; (3) implicit 3D with only camera feature; (4) with both image and camera features (ours). 
We evaluate our model on scene consistency. Results in~\Cref{tab:ablation_vggt} validate our claim that the fusion of scene’s content information and camera-viewpoint information effectively creates a rich and joint representation for video generation in scene consistency.

\begin{table}[t]
\vspace{-10pt}
	\centering
	\caption{\textbf{Ablation on scene implicit 3D representation}.}
	\label{tab:ablation_vggt}
	\vspace{-2mm}
	\setlength{\tabcolsep}{4.5pt}
	\footnotesize
        \resizebox{\linewidth}{!}{%

\begin{tabular}{lccccc}
\toprule
& Mat. Pix.(K)$\uparrow$ & CLIP-V$\uparrow$ & PSNR$\uparrow$ & SSIM$\uparrow$ & LPIPS$\downarrow$ \\
\midrule
Loss-Guided &4509.46	&0.8552	&13.9996	&0.4020	&0.4458 \\
\midrule
W/o Implicit & 4527.46 & 0.8456 & 13.7582 & 0.3997 & 0.4506 \\
W/ Image feature & 4519.11 & 0.8518 & 13.9623 & 0.3981 & 0.4474 \\
W/ Camera feature & 4498.83 & 0.8544 & 14.1058 & 0.3997 & 0.4467 \\
\midrule
\textbf{Ours} & \textbf{4617.51} & \textbf{0.8633} & \textbf{14.5094} &\textbf{ 0.4133} & \textbf{0.4241} \\
\bottomrule
\end{tabular}
\vspace{-20pt}
        }
\end{table}

\begin{figure}[htbp]
    \centering
    \vspace{-10pt}
    \includegraphics[width=\linewidth]{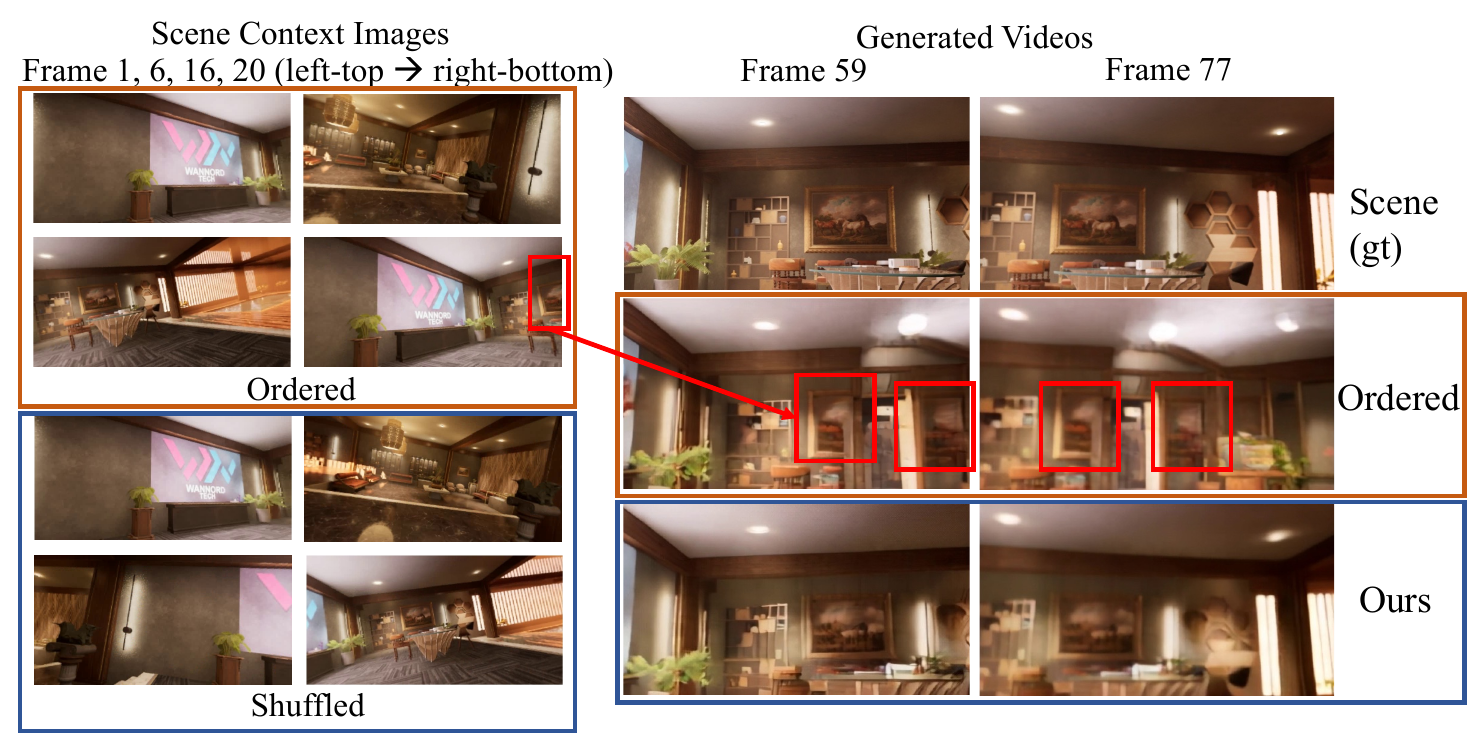}
    \vspace{-20pt}
    \caption{\textbf{Qualitative ablation study on shuffled context images.} The shuffled mechanism leads to better joint modeling and learning in scene consistency, while the ordered ones are tend to copy content from last provided image.
    }
    \vspace{-13pt}
    \label{fig:ablation_shuffle}
\end{figure}

\noindent
\textbf{Impact of Shuffled Context Images.}
We further analyze the impact of using shuffled mechanism for scene context images. We compare with the ordered scene context images and progressive training described in~\Cref{method: shuffled}. 
We evaluate our model on scene consistency and camera accuracy. \Cref{tab:ablation_shuffle} shows that the shuffled mechanism leads to better joint modeling and learning in scene consistency and camera accuracy. 
We show the qualitative comparison in~\Cref{fig:ablation_shuffle}, that the ordered scene context will let the model copy the content of the last image (marked in red square, copy twice in the generated video), which is inconsistent with the scene, neglecting the learning from implicit 3D.

\begin{table}[t]
    \vspace{-5pt}
	\centering
	\caption{\textbf{Ablation on shuffled context images}.}
	\label{tab:ablation_shuffle}
	\vspace{-2mm}
	\setlength{\tabcolsep}{4.5pt}
	\footnotesize
        \resizebox{\linewidth}{!}{%

\begin{tabular}{lcccccccc}
\toprule
& Mat. Pix.(K)$\uparrow$ & CLIP-V$\uparrow$ & PSNR$\uparrow$ & SSIM$\uparrow$ & LPIPS$\downarrow$ & RotErr$\downarrow$ & TransErr$\downarrow$ & CamMC$\downarrow$ \\
\midrule
Ordered & \textbf{4673.67} & 0.8592 & 14.0174 & 0.4099 & 0.4316 & 2.7329 & 5.1756 & 6.9152 \\
Progressive & 4531.66 & 0.8608& 14.2008 & 0.4111 & 0.4378 &\textbf{2.5757}	& 5.4059	& 7.0612  \\
\textbf{Shuffled (Ours)} & 4617.51 & \textbf{0.8633} & \textbf{14.5094} & \textbf{0.4133} & \textbf{0.4241} & 2.6825 & \textbf{5.146} & \textbf{6.8819} \\
\bottomrule
\end{tabular}

        }
\vspace{-20pt}
\end{table}

\subsection{Applications of \name{}}
\label{exp: generalization}
In this section, we show qualitative out-of-domain results in self-constructed samples from DiT360, and show potential applications in cinematic video generation.

\begin{figure}[t]
    \centering
    \vspace{-10pt}
    \includegraphics[width=\linewidth]{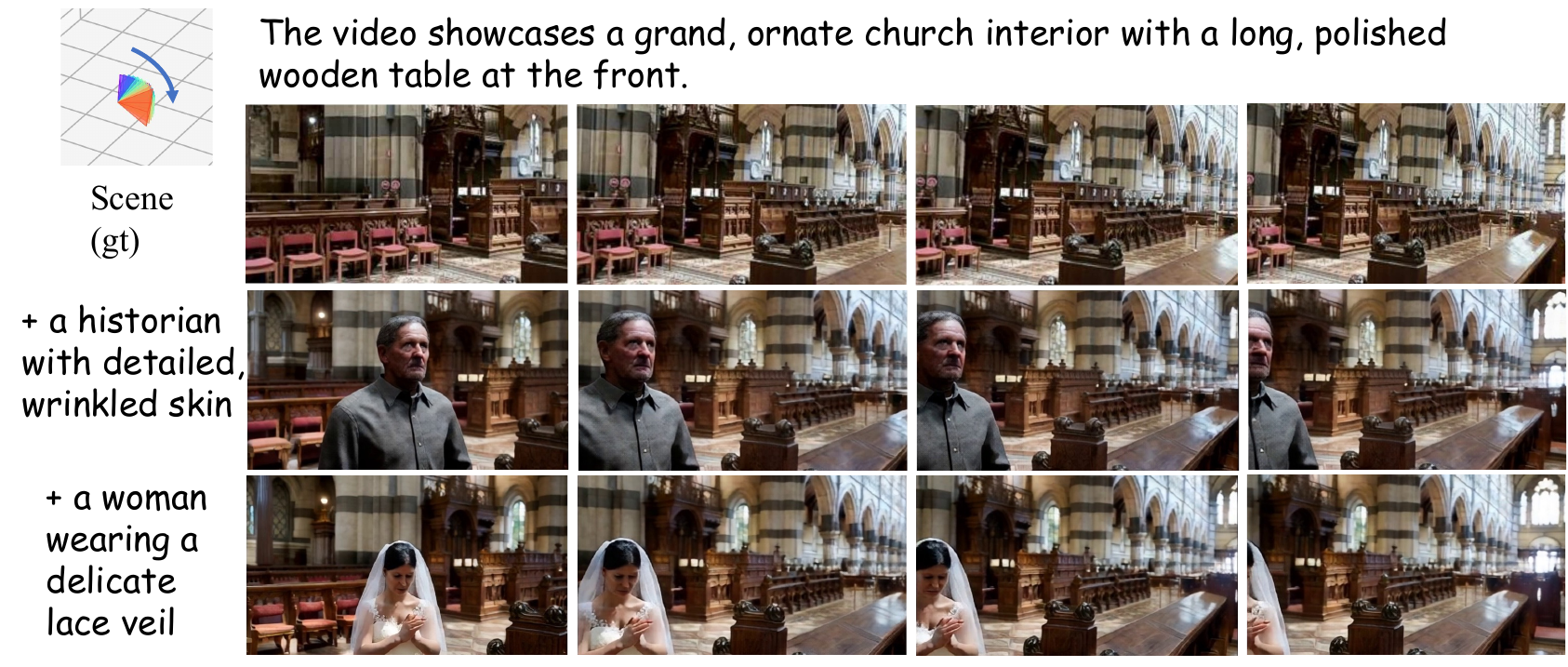}
    \vspace{-20pt}
    \caption{
    \textbf{Qualitative results of \name{} with out-of-domain results.} Our method has the potential to generalize to real-world scenarios.
    }
    \vspace{-10pt}
    \label{fig:ood_qualitative}
\end{figure}

\noindent
\textbf{Out-of-Domain Results.}
Shown in~\Cref{fig:ood_qualitative}, our method has the potential to generalize to real-world scenarios.

\begin{figure}[t]
    \centering
    \includegraphics[width=\linewidth]{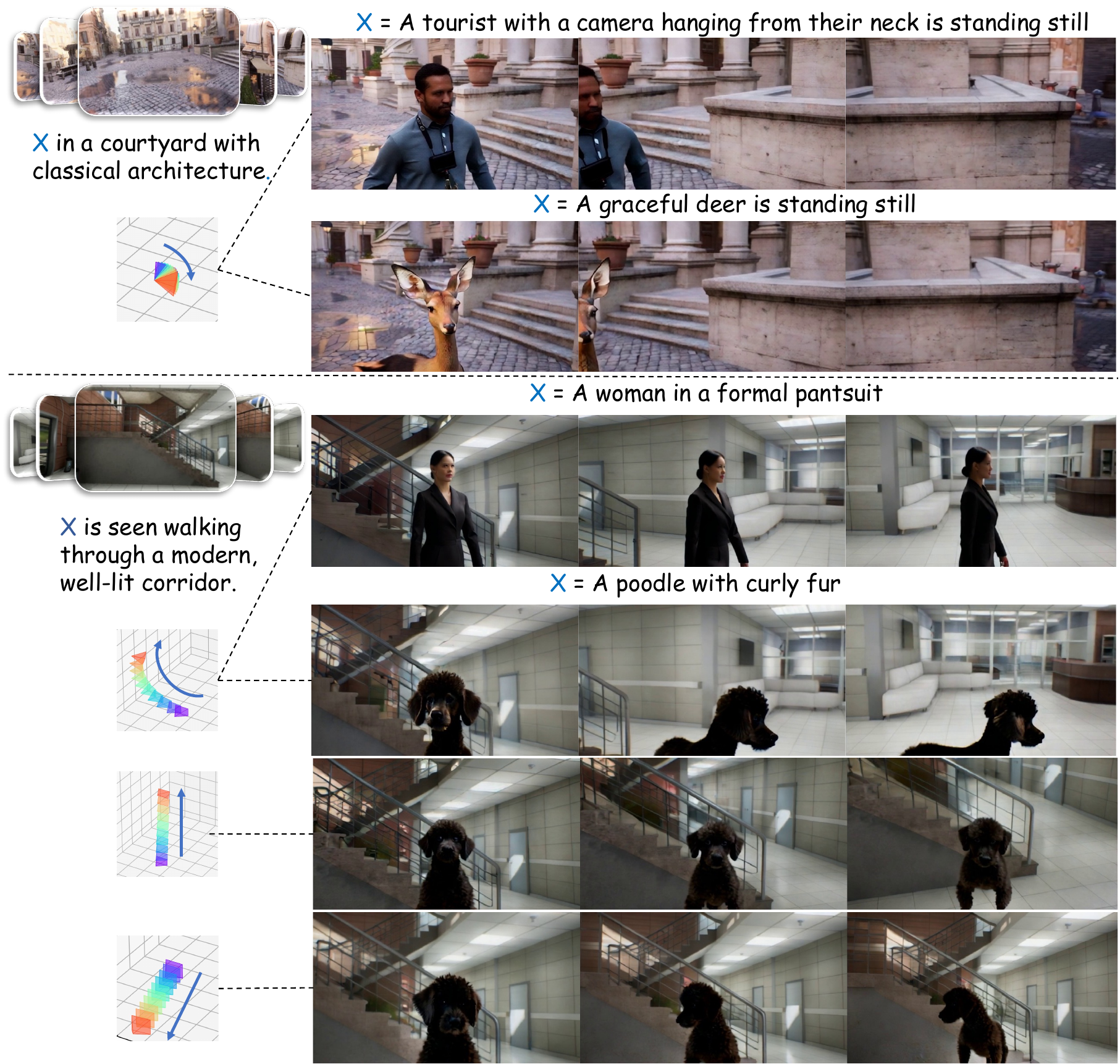}
    \vspace{-20pt}
    \caption{\textbf{Qualitative results of \name{} with diverse scenes, dynamic subjects, and camera trajectories.} Our method shows promising application of virtual stage and cinematic language in cinematic video generation.}
    \vspace{-15pt}
    \label{fig:application}
\end{figure}

\noindent
\textbf{Application of Virtual Stage.} Shown in~\Cref{fig:application} (first 4 rows), \name{} facilitates the creation of diverse performances by different subjects within a consistent virtual environment, streamlining production workflows.

\noindent
\textbf{Application of Cinematic Language.} Shown in~\Cref{fig:application} (last two rows), \name{} empowers creators to explore diverse camera trajectories for a given performance and static scene, allowing them to shape narrative impact through dynamic camera work, with scene consistency.

\section{Conclusion}
We prensent~\name{}, a cinematic video generation framework that can generate scene-consistent, camera-controlled videos with dynamic subject based on a decoupled scene representation. The method eliminates the need for explicit 3D information while keeping scene consistency in large view changes.
We explore the method of injecting implicit 3D-aware scene representation into pretrained T2V model, leveraging its generative capabilities to create dynamic videos. 
We proposed a simple yet effective shuffling mechanism for the input scene images during both feature extraction and concatenation. Furthermore, we developed a scene-decoupled video dataset using Unreal Engine 5, featuring diverse dynamic content, camera movements, and a variety of static scenes. 
Extensive experiments demonstrated \name{}’s state-of-the-art performance and promising applications.

{
    \small
    \bibliographystyle{ieeenat_fullname}
    \bibliography{ref}
}

\clearpage

\setcounter{page}{1}
\maketitlesupplementary

\appendix

\section{Dataset Construction}
\label{app: dataset}
We follow~\cite{bai2025recammaster} to use Unreal Engine 5 to construct Scene-Decoupled Video Dataset as described in~\Cref{sec: dataset}. 

\noindent
\textbf{3D Environments.} We collect 35 different 3D environment assets from \href{https://www.fab.com/}{here}. We mitigate the domain shift from rendered data to real-world distributions by prioritizing high-fidelity 3D assets while incorporating surreal environments as an auxiliary. Environmental variety is achieved through a diverse selection of internal and external venues, ranging from metropolitan streets and commercial interiors to pastoral settings.

\noindent
\textbf{Subjects.} We collect 70 different human 3D models as subjects from \href{https://www.fab.com/}{source A} and \href{https://www.mixamo.com/#/}{source B}, covering a wide range of styles such as realistic, anime, game-style.

\noindent
\textbf{Animations.} We collect around 100 different animations from \href{https://www.fab.com/}{source A} and \href{https://www.mixamo.com/#/}{source B}, including  diverse dynamic animations (\textit{e.g.}, waving, dancing, cheering.

\noindent
\textbf{Camera Trajectories}
In order to achieve large view changes, 
we add constraints to the camera movements.  The speed of camera movements is set to be linear.

\begin{enumerate}

\item \textbf{Arc Movements (Left/Right)}: For ``arc'' movements, the camera traverses an arc spanning 75 degrees over 77 frames, continuously tracking the dynamic subject to ensure it remains in view.

\item 
\textbf{Pan Movements (Left/Right)}: For ``pan'' movements, we require a 75-degree rotation over 77 frames. This movement involves only camera rotation, with no change in its spatial position.

\item 
\textbf{Arc and Tilt Movements (Up/Down)}: The vertical camera motion is constrained between 10 and 45 degrees relative to its initial orientation. For ``arc'' movements, the camera tracks the dynamic subject. For ``tilt'' movements, the camera only rotates with no change in translation.

\item 
\textbf{Dolly, Truck, and Pedestal Movements (Left/Right/Up/Down/Forward/Backward)}: For dolly, truck, and pedestal movements, we randomly select the distance thresholds relative to the dynamic subject. The specific ranges for these movements are defined as follows:
\begin{itemize}
    \item Left: $[1/4, 2]$
    \item Right: $[1/4, 2]$
    \item Forward: $[1/4, 5/4]$
    \item Backward: $[1/4, 2]$
    \item Up: $[1/4, 2/3]$
    \item Down: $[1/4, 2/3]$
\end{itemize}
These values represent the relative distance changes with respect to the subject.
\end{enumerate}

\noindent
\textbf{Initial Viewpoint}: The starting viewpoint for all camera movements and the panoramic images is randomly selected within a range of -45 to 45 degrees. This angle is measured relative to the front-facing direction of the dynamic subject, introducing variability in initial perspectives.

\section{Implementation Details}
Our model's trainable components include: 1) the 3D attention, projector, and feedforward layers within the DiT Blocks; 2) a convolutional layer and a Layer Normalization module for projecting the implicit 3D features; and 3) a learnable camera encoder for projecting the camera condition~\cite{bai2025recammaster}.
For the baselines compared~\cite{bai2025recammaster, xiao2025trajectory, ren2025gen3c}, we utilize the default parameters (\textit{e.g.}, number of generated frames) provided in their official GitHub repositories to ensure a fair evaluation of their standard performance.

\section{Experimental Results}

\subsection{Out-of-Domain Test}
To evaluate the model's generalization capabilities, we conduct an out-of-domain (OOD) test using the DiT360 dataset~\cite{feng2025dit360, Matterport3D}. This section first explains the construction of our test set and then presents the quantitative and qualitative results.

\noindent
\textbf{OOD Test Set Construction.}
The DiT360 dataset~\cite{feng2025dit360, Matterport3D} comprises static panoramic views stitched from RGB-D images of building-scale scenes. From this dataset, we selected 50 panoramic images as our test samples. Since each panorama originates from a single viewpoint, we are limited to ``pan'' camera trajectories, which involve only rotational movements without any translation.
To construct the test set, we process each panoramic image in two ways:
1) We project the panorama into a series of context images by sampling viewpoints along the horizontal plane at $18^\circ$ increments;
2) We define three distinct camera trajectories to generate the ground truth sequences.
It is important to note that this test set exclusively contains static scenes. Consequently, this evaluation focuses on comparing the methods' abilities to generate coherent and consistent static environments.

\noindent
\textbf{Quantitative Results on OOD Test Set.}
\begin{table*}[htbp] %
    \centering
    \caption{\textbf{Quantitative comparision with previous methods on OOD test set.} We compare \name{} with FramePack~\cite{zhang2025framepack} on scene consistency, Context-as-Memory~\cite{yu2025contextasmemory} and Gen3C~\cite{ren2025gen3c} on both scene consistency and camera accuracy, Traj-Attn~\cite{xiao2025trajectory} and RecamMaster~\cite{bai2025recammaster} on camera accuracy. We follow~\cite{xie2025simplegvr} to evaluate video quality on VBench.
    }
    \label{tab: app_compare_baselines_ood} %
    \adjustbox{max width=\textwidth}{
    \begin{tabular}{l ccccc ccc c c} 
        \toprule
        \multirow{2}{*}{\textbf{Model}} & \multicolumn{5}{c}{\textbf{Scene Consistency}} & \multicolumn{3}{c}{\textbf{Camera Accuracy}} & {\textbf{Text Alignment}}  & {\textbf{Video Quality}}
        \\
        \cmidrule(lr){2-6} \cmidrule(lr){7-9} \cmidrule(lr){10-10} \cmidrule(lr){11-11}
        & Mat. Pix.(K)$\uparrow$  & CLIP-V$\uparrow$ & PSNR$\uparrow$ & SSIM$\uparrow$ & LPIPS$\downarrow$ & RotErr$\downarrow$ & TransErr$\downarrow$ & CamMC$\downarrow$ & CLIP-T$\uparrow$  & VBench$\uparrow$ \\
        \midrule
        \textbf{Context-based Method} \\ %
        \midrule
        FramePack~\cite{zhang2025framepack} & 4025.98 & 0.8784 & 9.9902 & 0.3529 & 0.6625 & - & - & - & \textbf{0.3223} & 0.7813\\
        Context-as-Memory~\cite{yu2025contextasmemory} & 4338.75 & 0.8976 & 10.1676 & 0.364 & 0.6698 & 5.5482 & 11.2567 & 14.0722 & 0.3112 & \textbf{0.8093}\\
        \midrule
        \textbf{Explicit 3D Guidance Method} \\ %
        \midrule
        Gen3C~\cite{ren2025gen3c} & 4246.33 & \textbf{0.9049} & 11.9062 & 0.4412 & 0.6309 & 4.2975 & 10.8731 & 13.1951 & 0.3004 & 0.738\\
        \midrule
        \textbf{Camera-Controlled Method} \\ %
        \midrule
        Traj-Attn~\cite{xiao2025trajectory} & - & - & - & - & - & 4.3117 & 13.2458 & 14.899 & 0.3072 & 0.7635\\
        RecamMaster~\cite{bai2025recammaster} & - & - & - & - & - & 3.7125 & 11.3909 & 13.0711 & 0.3052 & 0.7705\\
        \midrule
        \textbf{Ours} & \textbf{4726.57} & 0.9038 & \textbf{12.0215} & \textbf{0.4470} & \textbf{0.5121} & \textbf{3.6981} & \textbf{10.8435} & \textbf{12.3307} & 0.3025 & 0.7965\\ %
        \bottomrule
    \end{tabular}
    }
\end{table*}

\begin{table*}[htbp]
    \centering
    \caption{\textbf{Ablation study on number of scene context images.} }
    \label{tab:app_ablation_number} %
    \resizebox{0.7\linewidth}{!}{
    \begin{tabular}{l ccccc ccc}
        \toprule
        \multirow{2}{*}{\textbf{\# Num}} & \multicolumn{5}{c}{\textbf{Scene Consistency}} & \multicolumn{3}{c}{\textbf{Camera Accuracy}} \\
        \cmidrule(lr){2-6} \cmidrule(lr){7-9}
        & Mat. Pix. (K)$\uparrow$ & CLIP-V$\uparrow$ & PSNR$\uparrow$ & SSIM$\uparrow$ & LPIPS$\downarrow$ & RotErr$\downarrow$ & TransErr$\downarrow$ & CamMC$\downarrow$ \\
        \midrule
        1 & 3614.14 & 0.7876 & 10.3426 & 0.1682 & 0.6478 & 3.9551 & 8.7780 & 10.8669 \\
        4 & 4200.82	&0.8167	&10.7466	&0.2215	&0.6069	&\textbf{2.3948}	&7.7557	&8.9438\\
        10 & 4519.10 & 0.8389 & 11.9169 & 0.2890 & 0.5447 & 3.0033 & 7.9496 & 9.5312 \\
        20 & \textbf{4617.51} & \textbf{0.8633} & \textbf{14.5094} & \textbf{0.4133} & \textbf{0.4241} & 2.6825 & \textbf{5.1460} & \textbf{6.8819} \\
        \bottomrule
    \end{tabular}
    }
\end{table*}

\begin{table}[thbp]
	\centering
	\caption{\textbf{Ablation on camera control condition}.} %
	\label{tab:ablation_camera} %
	\vspace{-2mm}
	\setlength{\tabcolsep}{4.5pt}
	\footnotesize
        \resizebox{0.6\linewidth}{!}{%

\begin{tabular}{lccc}
\toprule
& RotErr$\downarrow$ & TransErr$\downarrow$ & CamMC$\downarrow$ \\
\midrule
W/o Implicit & 2.7362 & 5.4411 & 7.1805 \\
W/o Camera & 11.3678	& 11.1976	& 19.5482 \\
\textbf{Ours} & \textbf{2.6825} & \textbf{5.146} & \textbf{6.8819} \\
\bottomrule
\end{tabular}

        }
\end{table}

\begin{table*}[t]
    \centering
    \caption{\textbf{Ablation study on scene descriptions.} }
    \label{tab:ablation_scene_description}
    \resizebox{1.0\linewidth}{!}{
    \begin{tabular}{l ccccc ccc c c}
        \toprule
        \multirow{2}{*}{\textbf{Method}} & \multicolumn{5}{c}{\textbf{Scene Consistency}} & \multicolumn{3}{c}{\textbf{Camera Accuracy}} & \textbf{Text Alignment} & \textbf{Video Quality} \\
        \cmidrule(lr){2-6} \cmidrule(lr){7-9} \cmidrule(lr){10-10} \cmidrule(lr){11-11}
        & Mat. Pix.(K)$\uparrow$ & CLIP-V$\uparrow$ & PSNR$\uparrow$ & SSIM$\uparrow$ & LPIPS$\downarrow$ & RotErr$\downarrow$ & TransErr$\downarrow$ & CamMC$\downarrow$ & CLIP-T$\uparrow$ & VBench $\uparrow$ \\
        \midrule
        remove scene description      & 4589.18 & 0.8576 & 14.3361 & \textbf{0.4159} & 0.4293 & 2.4974 & 5.3 & 6.8412 & 0.3057 & 0.8022 \\
        add inconsistent description   & 4577.54 & 0.8547 & 14.2669 & 0.4103          & 0.4308 & \textbf{2.474} & 5.2749 & \textbf{6.7745} & 0.3048 & 0.8006 \\
        \midrule
        \textbf{Ours}                         & \textbf{4617.51} & \textbf{0.8633} & \textbf{14.5094} & 0.4133 & \textbf{0.4241} & 2.6825 & \textbf{5.146} & 6.8819 & \textbf{0.3212} & \textbf{0.8053} \\
        \bottomrule
    \end{tabular}
    }
\end{table*}

\begin{table*}[t]
    \centering
    \caption{\textbf{Ablation study on different feature fusion strategies.} }
    \label{tab:finegrained_ablation}
    \resizebox{1.0\linewidth}{!}{
    \begin{tabular}{l ccccc ccc c c}
        \toprule
        \multirow{2}{*}{\textbf{Method}} & \multicolumn{5}{c}{\textbf{Scene Consistency}} & \multicolumn{3}{c}{\textbf{Camera Accuracy}} & \textbf{Text Alignment} & \textbf{Video Quality} \\
        \cmidrule(lr){2-6} \cmidrule(lr){7-9} \cmidrule(lr){10-10} \cmidrule(lr){11-11}
        & Mat. Pix.(K)$\uparrow$ & CLIP-V$\uparrow$ & PSNR$\uparrow$ & SSIM$\uparrow$ & LPIPS$\downarrow$ & RotErr$\downarrow$ & TransErr$\downarrow$ & CamMC$\downarrow$ & CLIP-T$\uparrow$ & VBench $\uparrow$ \\
        \midrule
        channel-dim        & 4564.53 & 0.8301 & 14.0434 & 0.4024 & 0.4372 & 2.4813 & 5.0500 & 6.6230 & 0.3170 & 0.8041 \\
        view-dim           & 4407.87 & 0.8265 & 12.8336 & 0.3757 & 0.4974 & 2.4915 & 5.9913 & 7.4144 & 0.3213 & 0.7982 \\
        separate $F_i, F_c$ & 4546.81 & 0.8533 & 14.1516 & 0.4037 & 0.4449 & 2.7345 & 5.4331 & 7.1764 & \textbf{0.3229} & 0.8022 \\
        \midrule
        \textbf{Ours}               & \textbf{4617.51} & \textbf{0.8633} & \textbf{14.5094} & \textbf{0.4133} & \textbf{0.4241} & 2.6825 & 5.1460 & 6.8819 & 0.3212 & \textbf{0.8053} \\
        \bottomrule
    \end{tabular}
    }
\end{table*}

\begin{table}[th]
    \centering
    \caption{Ablation on camera control condition. Supplement to~\Cref{tab:ablation_vggt}} 
    \label{tab:ablation_vggt_camera}
    \setlength{\tabcolsep}{6pt}
    \footnotesize
    \resizebox{0.8\linewidth}{!}{%
    \begin{tabular}{lccc}
        \toprule
         & RotErr$\downarrow$ & TransErr$\downarrow$ & CamMC$\downarrow$ \\
        \midrule
        w/o Implicit        & 2.7362 & 5.4411 & 7.1805 \\
        w/ Image feature    & 2.4967 & 5.2081 & 6.7754 \\
        w/ Camera feature   & 2.7631 & 5.1620 & 6.9542 \\
        \midrule
        w Implicit (Ours) & 2.6825 & 5.146 & 6.8819 \\
        \bottomrule
    \end{tabular}
    }
\end{table}

\begin{table}[htbp]
    \centering
    \caption{\textbf{Ablation on OOD trajectories.}} 
    \label{tab:ablation_ood_traj}
    \resizebox{\linewidth}{!}{
    \begin{tabular}{l ccc c c}
        \toprule
        \multirow{2}{*}{\textbf{Method}} & \multicolumn{3}{c}{\textbf{Camera Accuracy}} & \textbf{Text Alignment} & \textbf{Video Quality} \\
        \cmidrule(lr){2-4} \cmidrule(lr){5-5} \cmidrule(lr){6-6}
        & RotErr$\downarrow$ & TransErr$\downarrow$ & CamMC$\downarrow$ & CLIP-T$\uparrow$ & VBench$\uparrow$ \\
        \midrule
        ReCamMaster~\cite{bai2025recammaster} & \textbf{2.2485} & 7.4892 & 8.5504 & 0.3051 & 0.8015 \\
        Traj-Attn~\cite{xiao2025trajectory}   & 3.4825          & 8.6681 & 10.5595 & 0.2669 & 0.7728 \\
        \midrule
        \textbf{Ours }       & 2.4147          & \textbf{6.7794} & \textbf{8.0997} & \textbf{0.3163} & \textbf{0.8022} \\
        \bottomrule
    \end{tabular}
    }
\end{table}

The quantitative results are presented in~\Cref{tab: app_compare_baselines_ood}. Our proposed method demonstrates superior performance in scene consistency, as evidenced by its leading scores across the Mat. Pix., PSNR, SSIM, and LPIPS metrics. Furthermore, our model achieves the second-highest CLIP-V score. In terms of camera accuracy, the RotErr, TransErr, and CamMC metrics confirm the effectiveness of our method.

\noindent
\textbf{Qualitative Results on OOD Test Set.}
Please see the qualitative results in the supplementary video.

\subsection{Ablation Study}
\noindent
\textbf{Ablation on Number of Scene Context Images.}

We conduct four experiments on different numbers of scene context images: 1, 4, 10, and 20. 
The results are shown in~\Cref{tab:app_ablation_number}. 
The results show that increasing the number of scene context images leads to a consistent improvement in performance across both scene consistency and camera accuracy.
For scene consistency, all metrics show a clear positive trend. As the number of views increases from 1 to 20, the Mat. Pix. and CLIP-V scores steadily rise, while the LPIPS error consistently decreases. This demonstrates that more contextual information helps the model generate scenes with higher geometric fidelity and semantic coherence. PSNR and SSIM, show a marked increase, signifying a substantial enhancement in the clarity and structural integrity of the generated videos.
For camera accuracy, the overall trend also confirms the benefit of using more views, with the 20-view configuration achieving the lowest error rates across RotErr, TransErr, and CamMC. We observe a slight, non-monotonic degradation in performance when moving from 4 to 10 views. This might suggest that a moderate number of views can introduce temporary ambiguities for camera pose estimation, which are effectively resolved when a richer set of 20 views provides more robust constraints.

\noindent
\textbf{Impact of Camera Control.}
We conduct three experiments on camera accuracy: (1)  without implicit 3D information, with camera condition; (2) with implicit 3D information, without camera condition; (3) with both (ours). Results in~\Cref{tab:ablation_camera} show that the input camera instruction provides fine-grained and accurate control, while implicit 3D information further boosts the camera control.

\noindent
\textbf{Different Condition Mechanisms.}
We show the results in~\Cref{tab:finegrained_ablation}.
(1) Different concatenations.
We ablate injecting conditions on channel/view-dimension~\cite{bai2025recammaster}. We found that frame-dimension (ours) provides a more robust way for the spatio-temporal interactions among conditions, which is also observed in~\cite{bai2025recammaster}.
(2) Different Fusion Strategies.
We compare ours with concatenating $F_i$ and $F_c$ directly into tokens without element-wise addition. Our fusion strategy better integrates scene content and viewpoints, enhancing consistency.
(3) The camera accuracy are on par among the different mechanisms, as it is mainly impacted by injecting camera trajectory (also validated in~\Cref{tab:ablation_camera} and~\Cref{tab:ablation_vggt_camera}).

\noindent
\textbf{Inconsistent Modalities.}
We evaluate robustness by removing scene descriptions or using inconsistent prompts (e.g., ``in the garden''). Despite a slight performance drop, our method remains robust to missing or mismatched descriptions, shown in~\Cref{tab:ablation_scene_description}.

\noindent  
\textbf{Out-of-range trajectories.}
Following~\cite{luo2025camclonemaster}, we evaluate 300 unseen trajectories from RealEstate10K~\cite{zhou2018stereo}. As groud truth videos are lack for consistency evaluation, we report all other available metrics in~\Cref{tab:ablation_ood_traj}. Our model shows generalization to out-of-range trajectories, benefiting from the 46K diverse trajectories in our dataset.

\subsection{More Qualitative Results}
Please refer to our \href{https://karine-huang.github.io/CineScene/}{project page}.

\section{Limitations}
Our current work presents several limitations that motivate future research directions:

1) We explore the generation of short video clips (77 frames) with a maximum view change of 75 degrees. Extending scene consistency to longer videos with larger view changes remains a challenging but important area for future investigation.

2) To simplify the problem, we provide the first scene context image with the same viewpoint as the first frame of the generated video. Future work will address the generation of videos from random camera positions.

3) \name{} inherits limitations from the pre-trained T2V models, such as distortion in humans' large motion movements.

\end{document}